# Code-Mixed Sentiment Analysis Using Machine Learning and Neural Network Approaches

**Pruthwik Mishra[1]   Prathyusha Danda[1]   Pranav Dhakras[2]**
Language Technologies Research Center, IIIT Hyderabad
Kohli Center On Intelligent Systems
{pruthwik.mishra, danda.prathyusha, pranav.dhakras}@research.iiit.ac.in

## Abstract

Sentiment Analysis for Indian Languages (SAIL)-Code Mixed tools contest aimed at identifying the sentence level sentiment polarity of the code-mixed dataset of Indian languages pairs (Hi-En, Ben-Hi-En). Hi-En dataset is henceforth referred to as HI-EN and Ben-Hi-En dataset as BN-EN respectively. For this, we submitted four models for sentiment analysis of code-mixed HI-EN and BN-EN datasets. The first model was an ensemble voting classifier consisting of three classifiers - linear SVM, logistic regression and random forests while the second one was a linear SVM. Both the models used TF-IDF feature vectors of character n-grams where n ranged from 2 to 6. We used scikit-learn (sklearn) (Pedregosa et al., 2011) machine learning library for implementing both the approaches. Run1 was obtained from the voting classifier and Run2 used the linear SVM model for producing the results. Out of the four submitted outputs Run2 outperformed Run1 in both the datasets. We finished first in the contest for both HI-EN with an F-score of 0.569 and BN-EN with an F-score of 0.526.

## 1 Introduction

Code-mixing refers to the use of linguistic units words, phrases, clauses from different languages at a sentence or utterance level. It is generally observed in an informal setting like social media. With the myriad social media platforms available for people to communicate, the quota of code-mixed data available to us is tremendous.

Code-mixed data can be easily extracted from various social-media platforms like Facebook, Twitter, web blogs etc. Even with the large volumes of data available to us, processing and analysis of code-mixed data still seems a herculean task. Sentiment analysis on social media data has become a popular research task in the recent years in the arena of text. Understanding the sentiment bound to a phrase or sentence would allow us to discern the opinion of the person. In case of a product review, the interests of the individual could be an asset to the businesses for advertising and marketing, that would help them in increasing customer satisfaction.

Considering the above points, the aim of Sentiment Analysis for Indian Languages (SAIL)-Code Mixed is to identify the sentence level sentiment polarity of the code-mixed dataset of Indian languages pairs (Hi-En, Ben-Hi-En). The data was collected from Twitter, Facebook, and WhatsApp. Each of the sentences is annotated with language information as well as polarity at the sentence level. The polarity of the sentence gives the sentiment of the sentence; 1, 0, -1 for positive, neutral and negative sentiments respectively.

To solve the above task, we implemented two models that use SVM(Cortes and Vapnik, 1995), logistic regression and random forests as classifiers. The first model was a ensemble voting classifier consisting of three classifiers, logistic regression and random forests along with SVM. The second model was a linear SVM. Both the models used TF-IDF(Sparck Jones, 1972) feature vectors of 2 to 6 character n-grams. We also experimented with the word level n-grams as features for both SVM and MLP as classifiers. The mean of all the GloVe(Pennington et al., 2014) vectors in a sentence (GloVe averaged) were also used as features for SVM and MLP. Bi-LSTM with GloVe was explored and the results for all the models on training data are reported in the paper.

The organization of the paper is as follows. Section 2 sheds some light on related work and section 3 gives details about the corpus. Different approaches employed are explained in the subsequent section along with features used. Evaluation is described in section 5. Results constitutes sections 6 and Error Analysis & Observation are presented in section 7. We conclude our paper with the Conclusion & Future Work section.

## 2 Related Work

The shared task SAIL-2015 (Patra et al., 2015) reported accuracies of different systems submitted on sentiment analysis for tweets in 3 major Indian languages Hindi, Bengali and Tamil. Different supervised algorithms like Naive Bayes, Support Vector Machines, Decision Trees were used for this task. The best performing system implemented Naive Bayes algorithm for classification of sentiments of tweets into one of the three classes - positive, negative or neutral.

(Sharma et al., 2015) used a lexicon-based approach for assigning overall sentiment score to Hindi-English code-mixed sentence. They used the FIRE 2013 (Roy et al., 2013) and FIRE 2014 datasets. (Joshi et al., 2016) used sub-word level representations in LSTM (Hochreiter and Schmidhuber, 1997) to better the state-of-art performance (Sharma et al., 2015) on the Hindi-English code-mixed datasets by 18%. The dataset consisted of the user comments from public Facebook pages of two of the most popular celebrities. (Hassan et al., 2016) used LSTMs for sentiment analysis of Bengali, they reported 78% accuracy on binary sentiment analysis and 55% on 3 class (positive, negative, neutral) sentiment classification.

## 3 Corpus Details

The corpus contained code-mixed sentences along with their corresponding sentiments. There were 3 sentiments, positive (+1), neutral (0), and negative (-1) with each sentences being assigned exactly one of these 3 sentiments. Additionally, the sentences were also language tagged, that is, each word was tagged with the corresponding language such as HI (Hindi), BN (Bengali), EN (English) and UN (Universal). An example from the Hindi-English training data is given below:-

"**text**": "i remember it happen in my place also koi apni jagah se nahin hilega",
"**lang_tagged_text**": "i\\EN remember\\EN it\\EN happen\\EN in\\EN my\\EN place\\EN also\\EN koi\\HI apni\\HI jagah\\HI se\\HI nahin\\HI hilega\\HI ",
"**id**": 18027,
"**sentiment**": -1

Sample Bengali-English sentences along with their sentiments are given below:-

- Sentence "Tamil movie guler # Nayok ghula jmn e hok na kno # STORY gula # awsm hoy kintu <3 : D" is labeled with a sentiment score 1.

- Sentence "Youtube theke Ghure Asun . Alokito Geani 2017 ar sponsored by Zee-Bangla" is tagged neutral and assigned a score 0.

- Sentence "I think abhinetra ektu , actually onektai overdone kore feleche dorkar er chaite ." is assigned a sentiment score of -1.

We used 85% of the data for training and rest 15% for validation. Table 1 lists the number of training, development and test samples. The test data was provided by the organizers.

| Lang  | # training | # dev | # test |
|-------|------------|-------|--------|
| Hi-En | 10995      | 1941  | 5525   |
| Bn-En | 2125       | 375   | 3038   |

Table 1: Number of samples in training, development and test corpora for Hi-En and Bn-En datasets

## 4 Approach

In this section, we give description of various models, the respective machine learning ap-proaches and the corresponding feature sets from the datasets provided.

| Model | Hyperparam | Value |
|---|---|---|
| MLP | # Units | 100 |
| | Dropout | 0.3 |
| | Optimizer | Adam |
| | Learning rate | 0.001 |
| Bi-LSTM | # Units | 100 |
| | Optimizer | RMSProp |
| | Learning rate | 0.01 |

Table 2: Model-wise hyperparameters

## 4.1 Features

### 4.1.1 TF-IDF

TF-IDF vectors have proven to be an effective mechanism to encode textual information into real-valued vectors. The TF-IDF package from sklearn (Pedregosa et al., 2011) library that we used, uses count vectorizers to convert text input into a collection of tokens. It gives the flexibility of including higher n-grams in the vocabulary. This can prove to be helpful in the classification task. We experimented with TF-IDF vectors for the words (unigrams), bi-grams and trigrams present in the training corpus. We also experimented with TF-IDF vectors for character n-grams with $n$ values varying from 2 to 6. We used different TF-IDF vectors for Hindi-English and Bengali-English corpora.

### 4.1.2 GloVe

Word embeddings have been popular in recent times to understand words as points in real-valued multidimensional vector space. Specifically we trained 300 dimensional GloVe vectors separately on both Hindi-English as well Bengali-English code-mixed data. We augmented the Hindi-English code-mixed sentences from the given corpus with code-mixed data from other sources to increase the robustness of the trained word embeddings. We use GloVe embeddings in two ways, one by averaging the word embeddings for all words in a given text and other by treating a given text as a sequence of word embeddings. While using word embeddings of any form, an important aspect is the handling of out of vocabulary words (OOVs). In both cases, we handle all OOVs by setting their word embeddings to zero. Active research is going on regarding how to handle OOVs with word embeddings including random estimates, fine tuning, leveraging character based embeddings etc. but we leave these techniques for purpose of future work.

## 4.2 Models

### 4.2.1 Support Vector Machines

Support Vector Machines (SVM) have long been popular and effective models for multi-class classification problems. We used Support Vector Machines(SVM) as one of the classifier and we experimented with different methods to generate feature vectors for the data. We used sklearn linear SVM library for implementing our SVM based models. We employed the one-versus-one strategy for the classification task.

## 4.3 Voting Classifier

We used an ensemble technique of voting classifier for the classification task. The estimators used were linear SVM, logistic regression and random forest. Soft voting technique was used to combine the predictions of the three classifiers. Soft voting predicts the class label based on the argmax of sum of predicted probabilities of the classes. We used this ensemble technique present in the sklearn library.

### 4.3.1 Neural Networks

With the advent of word embeddings, neural network approaches have yielded impressive results for various classification tasks. We implemented multi layer perceptron (MLP) (Lippmann, 1989) with one hidden layer in Keras (Chollet et al., 2015) and output layer with three units. We use softmax activation function and pick up the class with the highest probability as given by the softmax activation. We tried two feature sets with our MLP model, namely, unigram TF-IDF vectors and averaged GloVe vectors.

Recurrent Neural Networks and Long Short Term Memory Networks (Hochreiter and Schmidhuber, 1997) are a natural extension of neural networks for processing sequential (or partially sequential) data such natural language. We also implemented a Bi-LSTM (Graves and Schmidhuber, 2005) model with GloVe vectors as input to predict the sentiment of a given text. The LSTM layer acts as the hidden layer in this model. Table 2 shows the implementation details for neural network models including the number of units in hidden layer, dropout rate, optimizer and learning rate.

| Lang | Model | Features | Precision | Recall | F1-Score |
|---|---|---|---|---|---|
| Hi-En | SVM | uni | 0.55 | 0.54 | 0.55 |
| | | uni-bi | 0.58 | 0.57 | 0.57 |
| | | uni-bi-tri | 0.57 | 0.57 | 0.57 |
| | | char (2,6)gram | 0.59 | **0.58** | **0.58** |
| | | char (3,6)gram | 0.59 | 0.57 | 0.57 |
| | | GloVe avg | **0.60** | 0.54 | 0.55 |
| | MLP | uni | 0.48 | 0.48 | 0.48 |
| | | uni-bi | 0.54 | 0.52 | 0.53 |
| | | uni-bi-tri | 0.56 | 0.52 | 0.53 |
| | | GloVe avg | **0.60** | 0.52 | 0.53 |
| | Bi-LSTM | GloVe | 0.55 | 0.53 | 0.54 |
| | Voting Classifier | char(2,6)gram | 0.61 | 0.54 | 0.55 |
| Bn-En | SVM | uni | 0.68 | 0.65 | 0.66 |
| | | uni-bi | 0.69 | 0.64 | 0.66 |
| | | uni-bi-tri | 0.68 | 0.63 | 0.64 |
| | | char (2,6)gram | **0.71** | **0.69** | **0.69** |
| | | char (3,6)gram | **0.71** | 0.68 | **0.69** |
| | Voting Classifier | char(2,6)gram | 0.69 | 0.67 | 0.68 |

Table 3: Results on development data for Hindi-English, Bengali-English Dataset. Both word and character n-gram features use TF-IDF.

## 5 Evaluation

The evaluation[1] and validation scripts were provided by the organizers. The output of the system was required to be in json format, the validation code verified the validity of the submitted systems. Macro F1 score was used to evaluate every system. Macro F1 score of the overall system was the average of F1 scores of the individual classes. Here the classes involved were -1, 0 and 1 which denoted the negative, neutral, positive sentiments respectively.

## 6 Results

Table 3 shows the results on development data for Hindi-English and Bengali-English datasets. Table 4 shows the results on test data for Hindi-English and Bengali-English datasets. The best performing system measures have been marked in bold. In Table 3 we observed that two best performing models were:

- Voting Classifier
- SVM with (2,6) char n-grams as features

[1] https://github.com/brajagopalcse/SAIL_CodeMixed-ICON-2017

We submitted two models for testing the system the results of which are listed in Table 4.

## 7 Error Analysis & Observation

In the training data, there are some ambiguous samples. Some examples are detailed below.

- The hindi sentence "so pls" was annotated as a negative sample. But there is no clear indication for it to be negative.

- The hindi sentence "Acha acha sahi hai" is given a sentiment score -1 (negative). This sentence has positive sentiment words like "acha" and "sahi".

- The bengali sentence "I've never heard that before , do you mean ' toi manush ni na goru ?" is labeled as neutral. But the sample contains words indicating negative sentiments.

| Lang | Model | Parameter | Precision | Recall | F1-Score |
|---|---|---|---|---|---|
| Hi-En | Model 1 | Overall | 0.597 | **0.56** | **0.569** |
| | | Positive | 0.719 | 0.699 | 0.707 |
| | | Neutral | 0.674 | 0.671 | 0.663 |
| | | Negative | 0.691 | 0.644 | 0.659 |
| | Model 2 | Overall | **0.607** | 0.547 | 0.557 |
| | | Positive | 0.719 | 0.696 | 0.705 |
| | | Neutral | 0.669 | 0.661 | 0.649 |
| | | Negative | 0.709 | 0.628 | 0.646 |
| Bn-En | Model 1 | Overall | **0.552** | 0.531 | 0.524 |
| | | Positive | 0.67 | 0.64 | 0.641 |
| | | Neutral | 0.632 | 0.649 | 0.619 |
| | | Negative | 0.678 | 0.655 | 0.664 |
| | Model 2 | Overall | 0.551 | **0.534** | **0.526** |
| | | Positive | 0.664 | 0.633 | 0.633 |
| | | Neutral | 0.633 | 0.65 | 0.621 |
| | | Negative | 0.683 | 0.666 | 0.673 |

Table 4: Results on test data for Hindi-English and Bengali-English Dataset. Model 1 is voting classifier and Model 2 is linear SVM with TF-TDF features.

| Lang | Text | Exp | Pred |
|---|---|---|---|
| Hi-En | U have laptop with u. | 0 | -1 |
| | chalk churane ka bhi apna maza tha . . . :) | 1 | 0 |
| | bale doesnt celebrate his goal but races over cuddling dempsey on the 2nd . illustrates perfectly the bullshitting spotlight whores they are . | -1 | 1 |
| Bn-En | @ ColorsBangla Oindrila to Etodin tv theke baire 6lo bt tao oke newa hoy6e , o jokhn serial korto emn khub 1ta populr | 1 | -1 |
| | Suto boro shobai ke janai EID Mubarak and SALAM | 1 | -1 |

Table 5: Error analysis.

So the models learned on this ambiguous data also got confused on test data. We observed that char n-gram TF-IDF vectors outperform the word n-grams. This may be attributed to the agglutination of words in social media text. As there is a limit on the text size on some social media like twitter which has an upper limit of 140 characters, users tend to combine words. So word n-grams were not able to capture the sentiment of a sentence.

We could observe that the tokens in a code-mixed sentence have spelling variations . This in turn affects the classification accuracy. For example the word है has different romanized spelling variations like hai, haii, he, hey.

Some of the examples that were incorrectly predicted by our best model are shown in Table 5. The **Exp** column in the table denotes the expected sentiment as deemed by our human evaluators since the gold labels were not available to us. The **Pred** column in table 5 denotes the sentiment predicted by our systems.

The third example in Hindi-English maybe tagged incorrectly because of words like *celebrate*, *perfectly* and *spotlight* which usually have a positive sentiment. Similarly the sentiment in the second example in Bengali-English may be predicted incorrectly due to the low term frequencies of the tokens present in the text.

## 8 Conclusion & Future Work

In this paper, we showed that machine learning approaches and neural networks could achieve comparable accuracy in code-mixed social media sentiment analysis to the systems relying on hand-crafted features. The ensemble model performed well with limited amount of data.

In the future, we intend to use character embeddings along with the word embeddings to get better representation of a sentence and words. This will also help in getting a representation for out-of-vocabulary(OOV) words. We can explore multilingual embeddings for the languages used in the available text. This can also help improve the sentiment classification accuracy. We intend to include some linguistic regularization (Qian et al., 2016) while learning the bi-LSTM to take advantage of intensifiers, negative words, positive words and other cue words. We can also explore convolutional networks for better hierarchical representation for characters, words and sentences.

# References

François Chollet et al. 2015. Keras. https://github.com/fchollet/keras.

Corinna Cortes and Vladimir Vapnik. 1995. Support vector machine. *Machine learning*, 20(3):273–297.

Alex Graves and Jürgen Schmidhuber. 2005. Framewise phoneme classification with bidirectional lstm and other neural network architectures. *Neural Networks*, 18(5):602–610.

Asif Hassan, Nabeel Mohammed, and AKA Azad. 2016. Sentiment analysis on bangla and romanized bangla text (brbt) using deep recurrent models. *arXiv preprint arXiv:1610.00369*.

Sepp Hochreiter and Jürgen Schmidhuber. 1997. Long short-term memory. *Neural computation*, 9(8):1735–1780.

Aditya Joshi, Ameya Prabhu, Manish Shrivastava, and Vasudeva Varma. 2016. Towards subword level compositions for sentiment analysis of hindi-english code mixed text. In *COLING*, pages 2482–2491.

Richard P Lippmann. 1989. Pattern classification using neural networks. *IEEE communications magazine*, 27(11):47–50.

Braja Gopal Patra, Dipankar Das, Amitava Das, and Rajendra Prasath. 2015. Shared task on sentiment analysis in indian languages (sail) tweets-an overview. In *International Conference on Mining Intelligence and Knowledge Exploration*, pages 650–655. Springer.

Fabian Pedregosa, Gaël Varoquaux, Alexandre Gramfort, Vincent Michel, Bertrand Thirion, Olivier Grisel, Mathieu Blondel, Peter Prettenhofer, Ron Weiss, Vincent Dubourg, et al. 2011. Scikit-learn: Machine learning in python. *Journal of Machine Learning Research*, 12(Oct):2825–2830.

Jeffrey Pennington, Richard Socher, and Christopher Manning. 2014. Glove: Global vectors for word representation. In *Proceedings of the 2014 conference on empirical methods in natural language processing (EMNLP)*, pages 1532–1543.

Qiao Qian, Minlie Huang, and Xiaoyan Zhu. 2016. Linguistically regularized lstms for sentiment classification. *arXiv preprint arXiv:1611.03949*.

Rishiraj Saha Roy, Monojit Choudhury, Prasenjit Majumder, and Komal Agarwal. 2013. Overview of the fire 2013 track on transliterated search. In *Post-Proceedings of the 4th and 5th Workshops of the Forum for Information Retrieval Evaluation*, page 4. ACM.

Shashank Sharma, PYKL Srinivas, and Rakesh Chandra Balabantaray. 2015. Text normalization of code mix and sentiment analysis. In *Advances in Computing, Communications and Informatics (ICACCI), 2015 International Conference on*, pages 1468–1473. IEEE.

Karen Sparck Jones. 1972. A statistical interpretation of term specificity and its application in retrieval. *Journal of documentation*, 28(1):11–21.